\documentclass[conference, onecolumn]{IEEEtran}
\IEEEoverridecommandlockouts
\usepackage{cite}
\usepackage{amsmath,amssymb,amsfonts}
\usepackage{algorithmic}
\usepackage{graphicx, float}
\usepackage{textcomp}
\usepackage{xcolor}
\usepackage{graphics}

\usepackage{csquotes}
\usepackage{subcaption}
\usepackage{multicol}
\usepackage{lipsum}
\usepackage[bookmarks=false]{hyperref}

\usepackage{wrapfig}
\def\BibTeX{{\rm B\kern-.05em{\sc i\kern-.025em b}\kern-.08em
    T\kern-.1667em\lower.7ex\hbox{E}\kern-.125emX}}
    
\usepackage{tabularx}

\usepackage{algorithm}
\usepackage{authblk}
\usepackage{multirow}
\usepackage{booktabs}

\usepackage{tikz}
\usetikzlibrary{matrix}
\usepackage{xparse}
\usepackage{soul}
\usepackage{amsmath}

\usepackage[most]{tcolorbox}

\usepackage{float} % Add this package in the preamble
\newcommand*{\affaddr}[1]{#1} % No op here. Customize it for different styles.
\newcommand*{\affmark}[1][*]{\textsuperscript{#1}}
\newcommand*{\email}[1]{\textit{#1}}

\begin{document}

%\title{Neuro-Agentic Control: An Agentic AI Framework for Industrial IoT Security \\}
\title{Neuro-Agentic Control: A Deep Learning-based LLM-Powered Agentic AI Framework for Controlling Security Controls\\}

\author[]{}

\author{Saroj Gopali\affmark[1], Bipin Chhetri\affmark[1], Deepika Giri\affmark[2], Sima Siami-Namini\affmark[3], Akbar Siami Namin\affmark[1]\\
\affaddr{Department of Computer Science\affmark[1]}, 
\affaddr{Texas Tech University\affmark[1]}\\
\affaddr{Cumberland University\affmark[2]}, \affaddr{Advanced Academic Programs Science\affmark[3]}, 
\affaddr{Johns Hopkins University\affmark[3]}

\\
\email{\{saroj.gopali, bipin.chhetri, akbar.namin\IEEEauthorrefmark{1}\}}@ttu.edu, 
\email{dgiri25@students.cumberland.edu}, \email{ssiamin1@jhu.edu}
}

\maketitle

\begin{abstract}

Cyberattacks on operational technology are increasingly causing costly downtime and physical damage, exposing the limitations of traditional rule-based monitoring in industrial IoT environments. While Large Language Models (LLMs) have strong semantic reasoning abilities to assist in decision support, their hallucinatory nature presents unacceptable safety liabilities for closed-loop control. This paper introduces a neuro-agentic control framework, a novel architecture that couples an LLM-based planner (i.e., such as Gemini 2.5 Flash-Lite) with a pre-trained Time-Series Foundation Model (TimesFM), to achieve physics-grounded autonomous defense. The paper introduces a ``Counterfactual Physics Injection'' mechanism that simulates the impact of LLM-proposed interventions within the numerical latent space of the foundation model before actuation, while allowing the system to reject hallucinatory or unsafe actions. Evaluated on an industrial dataset (e.g., the Secure Water Treatment (SWaT)) in the context of stochastic attack scenarios, the framework exhibited better performance compared to LSTM and TCN baselines. The Neuro-Agentic Loop prevented five breaches (33.3\%) below the threshold versus LSTM (26.7\%) and TCN (13.3\%), with zero physically invalid (hallucinated) actions executed. These results demonstrate the efficacy of using foundation models as deterministic ``Sentinels'' to safeguard agentic AI in critical infrastructure.

\end{abstract}


\section{Introduction}

Cyberattacks on operational technology are increasingly causing real process disruptions and costly downtime for 
industrial organizations.  The Kaspersky ICS 2023 \cite{kaspersky_2023} identifies several instances in which manufacturers were forced to stop production lines due to ransomware and targeted intrusions. Meanwhile, 
analysis of unplanned downtime shows that manufacturers now lose more than \$2 million in the automotive 
sector for every hour lost, compared to approximately \$39,000 in fast-moving consumer goods \cite{smc_true_cost_downtime_2024}. In Cyber–Physical Systems (CPS), security failures are not abstract IT-related events, but immediate drivers of safety risk and large financial loss. In increasingly dense Industrial IoT deployments, billions of connected devices expand the attack surface, making manual device-by-device hardening and monitoring fundamentally infeasible at scale.

% Data-driven automation is becoming increasingly important for cyber-physical infrastructures and Industrial 
% Control Systems (ICS) in maintaining safety, dependability, and efficiency under complex dynamics and challenging 
% conditions \cite{xing2024security}. Industrial IoT and large-scale sensing deployments have created high-dimensional, 
% multivariate time-series streams where rule-based alarms and manual monitoring are insufficient for the early detection 
% and mitigation of critical events. Traditional engineering practice often relies on fixed thresholds, handcrafted interlocks, and human operator expertise, which do not scale well to fleets of heterogeneous assets and non-stationary environments. 

Large language models (LLMs) and time-series foundation models have recently become highly effective instruments for forecasting and 
reasoning tasks over high-dimensional streaming data. Together, LLMs and time-series models allow large-scale prediction for maintenance, decision support, and anomaly detection \cite{liang2024foundation,kevin2024anomaly}. Considering the surge of LLMs, foundation models present exciting possibilities for various applications as useful world models in safeguarding cyber-physical systems. Proactive risk mitigation is based on forecasting future trajectories in various operational scenarios. Recently, TimesFM and other time-series foundation models have been proposed as general-purpose forecasters trained in large and diverse time-series corpora~\cite{das2024decoder}.

% TimesFM uses a decoder-only transformer architecture. The model achieves higher accuracy with few-shot and zero-shot performance on a number of forecasting benchmarks, indicating that its latent representations capture rich information about system momentum, trends and seasonality \cite{li2025tsfm}. Meanwhile, LLM based agentic AI systems have demonstrated the potential to integrate experiments, improve code, and collaborate with tools in various fields such as IoT \cite{raheem2025agentic}, finance \cite{luqman2025agentic}, healthcare \cite{karunanayake2025next}, reasoning capacity\cite{karim2025transforming}, automation and time-series forecasting \cite{tiwari2025agentic} indicating that agentic workflows and concepts may be adopted for Industrial IoT environments \cite{pati2025agentic}. Although time-series-driven agents have demonstrated that LLMs can direct data selection, feature refinement, and model evaluation for large-scale forecasting tasks\cite{tiwari2025agentic}. These agents can query heterogeneous sensor streams, reason over device metadata, and initiate analytics or configuration routines in the context of the Internet of Things (IoT). 

% When explored collectively, this new joint application suggests that agentic LLM are perfectly suited to sit on top of time-series foundation models where they can transform high-level operational tasks into fine-grained semantic control actions. 

However, most existing agentic systems for IoT and time series forecasting operate in an offline setting and primarily focus on recommending or selecting models. The models utilized to refine code are limited to offline assistance and do not engage in closed-loop control of safety critical processes \cite{barenji2025agentic}. In addition, there is also recent evidence that foundation models can hallucinate \cite{chakraborty2025hallucination}, which raises questions about the reliability of such models to the design and context. An example will be hallucination in the medical field \cite{kim2025medical}, which could have been fatal.

% The introduced neuro-agentic framework is motivated by this gap. LLMs should be leveraged and function as semantic planners, suggesting structured semantic control actions, while a different physics-based module assesses these actions against a reliable base model of system dynamics prior to any actuation. 
This work introduces a Neuro-Agentic control framework that achieves physics-grounded, agentic control in industrial IoT settings by combining an LLM-based planner with a time-series foundation model. A central contribution of our work is the counterfactual physics injection mechanism, which embeds LLM-specified actions into the numerical input space of the time-series foundation model. Instead of updating or fine-tuning TimesFM's internal state, the framework uses a cumulative linear injection to perturb the recent historical window, simulating the impact of an action 
applied over a certain period of time. As a result, prior to any actual action being taken, the TimesFM Sentinel can function as a physics-grounded world model, predicting how any semantic control action will change the system trajectory. 

For the proposed work, we conducted experiments using the Secure Water Treatment (SWaT)\cite{mathur2016swat, 9671488} data set. The experiments focus on the non-linear tank-level variable LIT301 under multiple injected attack patterns. The Neuro-Agentic Loop is designed to autonomously identify new risks, develop counterfactual control strategies, and reject risky or hallucinatory actions. The key contributions of this paper are as follows:

\begin{enumerate}

     \item We propose a novel \textit{Neuro-Agentic Control} framework that explicitly separates semantic reasoning from physical control by coupling an LLM-based planner with a physics-grounded time-series foundation model for IoT systems.
    
    \item We introduce a \textit{Counterfactual Physics Injection} mechanism that translates LLM-generated semantic control actions into numerical perturbations of future forecasts. This enables deterministic validation of the proposed actions without finetuning the foundation model.
    
    \item We formulate a deterministic hallucination rejection strategy based on worst-case risk minimization, ensuring that unsafe or ineffective LLM-proposed actions are automatically rejected before execution.
    
    \item We demonstrate the effectiveness of the proposed framework on the SWaT dataset, showing improved breach prevention and risk reduction compared to LSTM- and TCN models in stochastic attack scenarios.
    
\end{enumerate}

The remainder of the paper is organized into the following sections. The related work is discussed in Section \ref{sec:related}. The technical overview of deep learning models is presented in Section \ref{sec:overview}. Section \ref{sec:methodology} contains the methodology and algorithm of the proposed framework. The experimental Section \ref{sec:experiment} contains information on the data set and deep learning architectures. Section \ref{sec:result} presents the results of the experiment. Section \ref{sec:discussion} presents the interpretability and limitations of Neuro-Agentic Loop. Section \ref{sec:conclusion} concludes the paper with future work.

\section{Related work}
\label{sec:related}

In order to more effectively incorporate exogenous variables into time series forecasting, TimeXer\cite{wang2024timexer} 
presents a modular transformer architecture.  TimeXer outperforms cutting-edge models like PatchTST, FEDformer, 
and Informer when tested on common benchmarks like ETT, Weather, and Electricity. It achieves up to 23.5\% 
improvement in MSE in long-term forecasting tasks. The findings demonstrate steady improvements in both univariate 
and multivariate contexts. However, TimeXer's more intricate architecture increases computational costs, which might make it 
less useful in settings with limited resources.

Qiu et al.  \cite{qiu2025dbloss} offer another technique, a decomposition-based loss function that uses Exponential 
Moving Averages (EMA) to model trend and seasonal components independently in order to improve time series forecasting. 
DBLoss consistently improves performance on eight real-world datasets, including ETT, Weather, Electricity, and Traffic, 
when integrated into models PatchTST\cite{nie2022time}, iTransformer\cite{liu2023itransformer} and foundation models GPT4TS\cite{zhou2023one} and CALF\cite{liu2025calf}. For instance, it reduces MSE from 0.439 to 0.423 
(PatchTST) on ETTh1 and from 0.361 to 0.344 on ETTm1, with comparable gains over forecasting horizons .
DBLoss also enhances few-shot fine-tuning and zero-shot generalization. However, the hyperparameters $\alpha$ (smoothing factor) and 
$\beta$ (seasonal-trend weighting), impacts its performance and must be adjusted for each data set and model.

Similarly, Kim et al.\cite{kim2025medical} tested 11 general-purpose and medical-specialized LLMs 
on seven hallucination-sensitive clinical reasoning and retrieval tasks in order to create a medical hallucination 
benchmark and evaluation framework using foundation models. According to their findings, general-purpose models like 
Gemini~2.5 Pro achieved up to 97\% hallucination-free responses with chain-of-thought(COT) prompting, while 
medical-specific models like MedGemma only achieved 28.6– 61.9\%, and even with mitigation techniques, hallucinations 
continued in the range 27–50\% of cases. 

Bhattacharjee et al.\cite{bhattacharjee2024zero} proposed a zero-shot LLM-guided counterfactual generation framework in which a large language model is prompted to generate semantically plausible counterfactual text examples for evaluating NLP 
classifiers without task-specific training data. However, it remains unclear how such zero-shot, LLM-driven 
counterfactual strategies would behave when coupled to real-time control of cyber-physical systems because Bhattacharjee et al.\cite{bhattacharjee2024zero}'s method concentrates on offline evaluation of static NLP models and lacks any physics-grounded or dynamical validation layer.

Ang et al.\cite{ang2025structured} similarly introduced TS-Agent, a modular LLM-based agentic framework that automates financial time series workflow. They empirically assessed TS-Agent on a variety of forecasting and synthetic generation tasks (such as stock, cryptocurrency, and volatility series). The author's techniques involve applying a planner agent with executable model code and a structured knowledge basis on prior model selection, code refinement, and hyperparameter tuning. However, the method is performed offline and on a code level optimization, which leaves an open question about the agent's behavior and if it could function better in safety critical situation.

\section{Technical Background}
\label{sec:overview}

\subsection{LSTM and TCN}
Long Short-Term Memory (LSTM)\cite{hochreiter1997long} is a Recurrent Neural Networks (RNNs) variant that uses gates in an architecture designed to address the vanishing  gradient problem. The fundamental architecture of an LSTM consists of a memory cell with input, output, and forget gates that control the information flow. Temporal Convolutional Networks (TCNs) are convolutional architectures for sequence modeling that employ causal dilated convolutions and residual connections to capture long-range temporal dependencies \cite{bai2018empirical}. 

\subsection{The Google TimesFM and Gemini 2.5 Flash lite}
Trained in a variety of large-scale chronological datasets \cite{das2024decoder},  TimesFM is a foundation model for time-series forecasting that leverages a transformer-based decoder-only architecture. A massive time-series corpus of 100 billion real-world data is used to pre-train TimesFM. Gemini 2.5  Flash-Lite \cite{gelman2025gemini25} \cite{gemini25flashlite_modelcard} is a sparse mixture-of-experts transformer that accepts text, image, audio, and video inputs with up to a 1M-token context and produces text outputs up to 64K  tokens, targeting high-volume, latency-sensitive applications like agentic workflows,  code assistance, and retrieval-augmented generation.

% Each of layers features multi-head self-attention (16 heads), feedforward sublayers, and 
% layer normalization. 

% Even with a relatively small size of 200M parameters, the model 
% demonstrates impressive zero-shot capabilities across unseen datasets with varied 
% domains and temporal granularities. The model can successfully capture both short- and 
% long-range dependencies with this configuration.

% \subsection{Gemini 2.5 Flash lite}
% Gemini 2.5 Flash-Lite \cite{gelman2025gemini25} is a lightweight member of Google’s 
% multimodal Gemini 2.5 family, designed to offer quick cost-efficient inference while 
% retaining strong reasoning and tool-use capabilities within a large context window.
% Flash Lite achieves inference speeds that are around 1.5 times faster than its 
% predecessor while maintaining competitive accuracy in time-series forecasting and 
% sophisticated reasoning tasks at far lower computational costs.
% According to the official model card\cite{gemini25flashlite_modelcard}, Gemini 2.5 
% Flash-Lite \cite{gelman2025gemini25} \cite{gemini25flashlite_modelcard} is a sparse mixture-of-experts transformer that accepts text, image, audio, 
% and video inputs with up to a 1M-token context and produces text outputs up to 64K 
% tokens, targeting high-volume, latency-sensitive applications  like agentic workflows, 
% code assistance, and retrieval-augmented generation.

\section{Methodology}
\label{sec:methodology}

\begin{figure*}[ht]
    \centering
    \includegraphics[width=\linewidth , height=4.5cm]{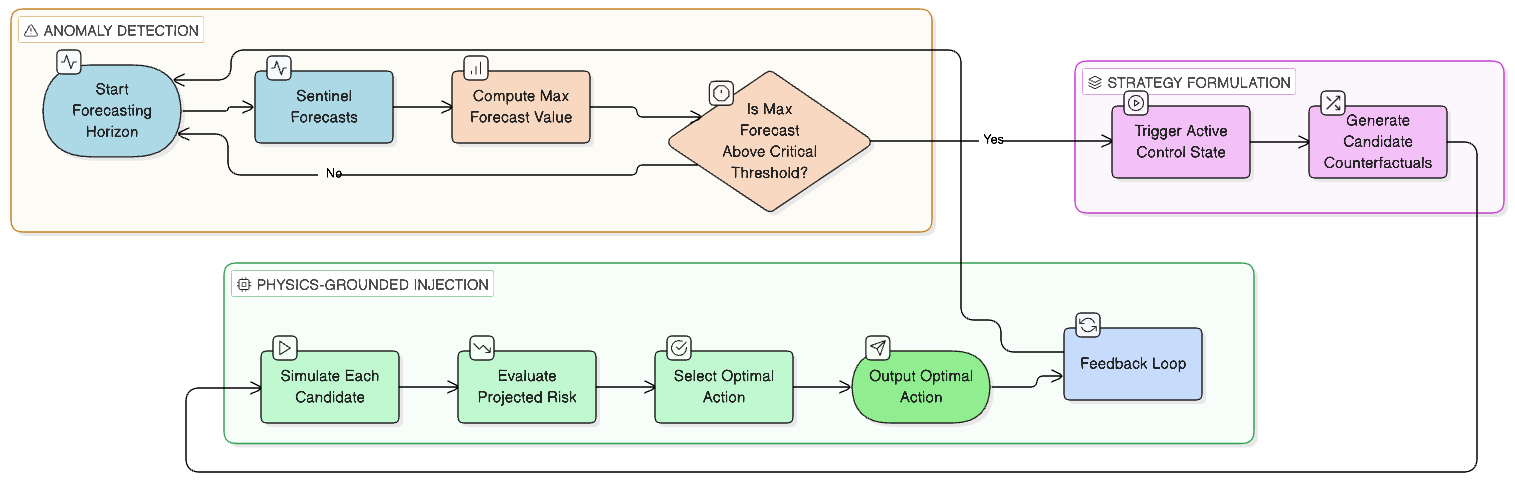}
    \caption{Neuro-Agentic System Flow Chart}
    \label{fig:neuro_flow}
    \vspace{-0.2cm}
\end{figure*}

The proposed architecture  \textbf{Neuro-Agentic Loop} Flow chart \ref{alg:neuro_agentic} operates as a two-stage control system. It integrates a semantic reasoner (i.e., The Architect) with a numerical physics engine (i.e., The Sentinel) to perform closed-loop controls on systems with multivariate time-series data. Algorithm \ref{alg:neuro_agentic} introduces the Neuro-Agentic Counterfactual Control Loop. It combines time-series prediction, retrieval-enhanced reasoning, and deterministic physics-based simulation to make safety-critical decisions. 
% The framework activates agentic intervention only when forecast risk exceeds a predefined threshold.  The actions are then systematically considered, with infeasible or unsafe strategies rejected before selecting the optimal control action.

\begin{algorithm}[t]%{!t}
    \caption{Neuro-Agentic Counterfactual Control Loop}
    \label{alg:neuro_agentic}
    \begin{algorithmic}[1]
    \small
        \REQUIRE 
            Historical window $X_t \in \mathbb{R}^{W}$, 
            Forecast horizon $H$, 
            Safety threshold $\theta$, 
            System Manual $M$
        \ENSURE Optimal Control Action $a^*$
        
        \STATE \textbf{Initialization:}
        \STATE $F \leftarrow \text{TimesFM (Sentinel)}$;   $G \leftarrow \text{Gemini-2.5-Flash (Architect)}$
        
        \STATE \textit{Step 1: Passive Sentinel Monitoring}
        \STATE $\hat{Y}_{base} \leftarrow F.\text{forecast}(X_t, \text{horizon}=H)$ ;  $R_{base} \leftarrow \max(\hat{Y}_{base})$
        
        \IF{$R_{base} < \theta$}
            \RETURN \text{"Monitor"} \COMMENT{System is safe}
        \ENDIF
        
        \STATE \textit{Step 2: Active Agentic Intervention}
        \STATE $\text{TargetID} \leftarrow \text{ExtractID}(X_t)$ 
        \STATE $\text{Context} \leftarrow \text{RetrieveRAG}(M, \text{TargetID})$
        \STATE $\text{Prompt} \leftarrow \text{Construct}(X_t, R_{base}, \theta, \text{Context})$
        
        \STATE \textit{\bf Architect proposes $k$ strategies}
        \STATE $S \leftarrow G.\text{generate\_strategies}(\text{Prompt})$ \COMMENT{$S = \{s_1, ..., s_k\}$}
        
        \STATE $a^* \leftarrow \text{"Monitor"}$; $R_{min} \leftarrow R_{base}$
        
        \FORALL{$s_i \in S$}
            \STATE \textit{Step 3: Feasibility Validation}
            \IF{$\neg \text{IsFeasible}(s_i)$}
                \STATE \textbf{continue} \COMMENT{Reject infeasible actions}
            \ENDIF
            
            \STATE \textit{Step 4: Counterfactual Physics Injection }
            \STATE $\hat{Y}'_{i} \leftarrow \hat{Y}_{base}$; $\mu, \delta \leftarrow s_i.\text{magnitude}, s_i.\text{duration}$
            
            \FOR{$t \leftarrow 0$ \TO $H-1$}
                \IF{$t < \delta$}
                    \STATE \textit{Cumulative drain effect during intervention}
                    \STATE $\hat{y}'_{t} \leftarrow \hat{y}_{t} + \mu \cdot (t + 1)$
                \ELSE
                    \STATE \textit{Maintain drained level after intervention ends}
                    \STATE $\hat{y}'_{t} \leftarrow \hat{y}_{t} + \mu \cdot \delta$
                \ENDIF
            \ENDFOR
            
            \STATE \textit{Step 5: Deterministic Risk Evaluation}
            \STATE $R_{sim} \leftarrow \max(\hat{Y}'_{i})$
            
            \STATE \textit{Step 6: Safety Verification}
            \IF{$R_{sim} < R_{min}$}
                \STATE $R_{min} \leftarrow R_{sim}$; $a^* \leftarrow s_i$
            \ELSE
                \STATE \textit{Reject strategy (hallucination or ineffective)}
            \ENDIF
        \ENDFOR
        
        \RETURN $a^*$
    \end{algorithmic}
 
\end{algorithm}

\begin{tcolorbox}[title=Neuro-Agentic Control Prompt, colback=gray!3, colframe=black!40]
\label{box:neuro_agentic_prompt}

\small

\textbf{System Role:} Autonomous Industrial Control Agent for the SWaT Water Treatment Plant.

\textbf{Context:} Operational manual and control constraints for the SWaT process.

\textbf{Current Situation:}

    Sensor: LIT301 (Tank Water Level);
     Current Level: $L_t$ mm;
    Predicted Peak Level: $\hat{L}_{\text{peak}}$ mm; 
    Critical Threshold: $L_{\text{crit}}$ mm;
    Excess Above Threshold: $\Delta L = \hat{L}_{\text{peak}} - L_{\text{crit}}$ mm;
    Time to Failure: $T_f$ time steps;
   Required Drain Rate: $R_{\text{req}}$ mm/timestep;
    Recent Trend: $\{L_{t-k}, \dots, L_t\}$.

\textbf{Task:} Generate three distinct intervention strategies to reduce the tank level below $L_{\text{crit}}$.

\textbf{Control Constraints:}

     \texttt{magnitude}: Negative real value (mm/timestep), representing drain rate;
  Typical operating ranges: 1)Emergency: $[-60, -40]$ (fast, high wear); 2) Moderate: $[-40, -20]$ (balanced); 3) Slow: $[-20, -10]$ (gentle, low wear);
 
    \texttt{duration}: Integer time steps in $[5, 50]$;
   Positive magnitudes (adding water) are forbidden;
     Total drained volume must satisfy: $|\text{magnitude} \times \text{duration}| \ge \Delta L$;

\textbf{Output Format:} JSON array only (no explanation):
\begin{verbatim}
[{"name": "...", 
"action_type": "trend_modification",
 "magnitude": <negative float>, 
 "duration": <integer> }]
\end{verbatim}

\end{tcolorbox}

\subsection{System Overview}
The system is modeled as a tuple $\{S, A, F\}$, where $S$ represents the state history of the time-series, $A$ is the set of potential semantic control actions, and $F$ is the prediction function. Unlike traditional control, where $F$ is a static differential equation, here $F$ is a pre-trained deep learning-based Foundation Model for time series prediction and forecasting (i.e., TimesFM). The workflow follows the following three stages:
(1) \textbf{Anomaly Detection.} The Sentinel continuously forecasts the horizon $H$. If $\max(F(x_t)) > \theta_{critical}$, where $x_t$ is the predicted time series element, the system triggers the active control state.
(2) \textbf{Strategy Formulation.} The Architect generates a set of $k$ candidate counterfactuals $C = \{c_1, ..., c_k\}$ using RAG-enhanced prompting. The RAG prompt includes system manuals, actuator restrictions, and safety thresholds. 
(3) \textbf{Physics-Grounded Selection.} The system simulates each counterfactual $c_i$ by applying interventions to the forecast and selects the optimal action $a^*$ with the objective function of minimizing the projected risk.

\subsection{Sentinel and Architect}
In this paper, Google's TimesFM\-2.5 with 200M parameters was utilized as the physics engine. As demonstrated \cite{gopali2025context}, TimesFM outperforms deep learning architectures (e.g., LSTM and TCN) in zero-shot forecasting. TimesFM achieves an RMSE of 0.3025 in the SWaT dataset. In this experiment, we freeze the model weights and utilize them purely for inference ($F(x)$), relying on its pre-trained knowledge of temporal dynamics to model system momentum and trend continuation.
The Architect \texttt{Gemini 2.5 Flash-Lite}, upon triggers, receives a prompt containing the recent sensor trend and the predicted time-to-failure (TTF) or anomaly. The LLM is constrained to output a structured JSON object that defines the control vector $c_i = \{ \text{action\_type}, \mu, \delta \}$, where $\mu$ represents the magnitude of the action (e.g., valve flow rate change) and $\delta$ represents the duration of the action in timesteps. All LLM outputs are validated to ensure $\mu < 0$ (for draining actions). The feasibility constraints are also checked ($|\mu| \leq 60$ mm/timestep, $5 \leq \delta \leq 50$ timesteps). The prompt is provided to  \texttt{Gemini 2.5 Flash-Lite} during the experiment, the contents of the prompt are demonstrated in a box \texttt{Neuro-Agentic Control Prompt} \ref{box:neuro_agentic_prompt}.

\begin{tcolorbox}[title=RAG Knowledge Base, colback=gray!3, colframe=black!40]
\small
% \textbf{Context:} LIT301: 
%     OPERATIONAL MANUAL SECTION 3.4 - TANK LEVEL CONTROL (LIT301):
%     \begin{itemize}
%         \item Normal Operating Range: 500mm to 800mm.
%         \item HIGH ALARM ($>$ 800mm): Immediate Risk of overflow.
%         \item CRITICAL LEVEL ($>$1000mm): Emergency intervention required.
%         \item INTERVENTION PROTOCOLS: 1) Open MV301 (Outflow/Permeate Valve) - Lowers level. 2) Stop P101/P102 (Feeding Pumps) - Stops rise immediately.
%         \item DANGER: Do not open MV201 during high alarm state without authorization.
%     \end{itemize}

\textbf{Manual \S3.4 -- Tank Level Control:} Normal range 500--800\,mm; HIGH ALARM $>$800\,mm (overflow risk); CRITICAL $>$1000\,mm (emergency intervention). \textbf{Intervention Protocols:} (1) Open MV301 outflow valve to lower level; (2) Stop feed pumps P101/P102 to halt rise. \textbf{Danger:} Do not open MV201 during HIGH ALARM without authorization.

\end{tcolorbox}

\subsection{Counterfactual Physics Injection}

The fundamental contribution of this work is the \textit{Physics Injection Function} $\Phi(\hat{Y}, c_i)$ that translates the semantic command $c_i$  into a numerical perturbation of TimesFM forecast ( the model's internal state cannot be directly modified). As a result, interventions must take place in the forecast. To estimate the refined forecast, the forecast must account for a recommended action with drain rate $\mu$ (negative) and duration $\delta$, the modified forecast $\hat{Y}'$ at time step $t$, which is denoted as follows:

\begin{equation}
    \hat{y}'_{t} = 
    \begin{cases}
        \hat{y}_{t} + \mu \cdot (t + 1) & \text{if } t < \delta \quad \text{(cumulative drain)} \\
        \hat{y}_{t} + \mu \cdot \delta & \text{if } t \geq \delta \quad \text{(maintain final level)}
    \end{cases}
\end{equation}

This equation defines an adjusted prediction $\hat{y}'_{t}$ that adds a time-dependent offset to the original prediction $\hat{y}_{t}$. For early time steps $t < \delta$, the offset increases linearly with time, representing a cumulative drain effect. After the threshold $t \geq \delta$, the offset is capped at $\mu \cdot \delta$, maintaining a constant final level. This formulation ensures that interventions affect the future forecast, not the immutable past. A drain valve opened at $t=0$ removes water cumulatively over time. After the valve closes at $t=\delta$, the water level remains at the drained state.

\subsection{Hallucination Rejection Algorithm}
\label{sec:halluc_def}

We distinguish two categories of rejected candidates. A hallucinated action is an LLM-proposed control vector that is physically invalid or unsafe in direction. This includes positive-magnitude (filling) under an overflow scenario, magnitude or duration outside actuator bounds, malformed JSON, or any vector whose counterfactual forecast exceeds the no-action baseline $R_{base}$ (i.e., it would actively worsen the state). An ineffective candidate is a well-formed, inbound vector that improves over baseline but is dominated by another candidate in the same set. Both are filtered, but only the former represents a safety-critical rejection. The final control decision is formulated as an optimization problem. The optimal action $a^*$ is: 

\begin{equation}
    a^* = \operatorname*{argmin}_{c \in C \cup \{ \emptyset \}} \left( \max_{t \in H} F(\Phi(\hat{Y}, c)) \right)
\end{equation}

% Where $\hat{Y}$ is the future forecast from the TimesFM baseline, $c$ is a candidate control action proposed by the LLM, ${\Phi(\hat{Y}, c)}$ is the counterfactual forecast after numerically injecting action and $F(⋅)$ is the TimesFM model. If the simulated risk of all proposed actions exceeds the risk of the null action ($\emptyset$), the system defaults to "Monitor," effectively rejecting the LLM's proposal as a hallucination. This physics-based validation is critical for safety in industrial on control systems.

Where $c$ is a candidate, $\Phi(\hat{Y}, c)$ is its counterfactual forecast, and $\emptyset$ is the null action. If every candidate's simulated risk exceeds the null risk, the system defaults to Monitor, treating the LLM proposals as hallucinated.

\section{Experimental Setup}
\label{sec:experiment}

\subsection{Dataset and Models Architecture}
We utilized the SWaT dataset (July version 2, 2019 dataset)\cite{itrustdatasets}, which contains information of an experimental water treatment that cybersecurity researchers utilize to design safe Cyber Physical Systems. The data contains $78$ columns with $14,996$ data points covering three hours of normal operation and one hour of attack incidents. In this experiment, we focus specifically on the \textit{Attack Incidents}, utilizing the target variable \texttt{LIT301} (i.e., Tank Level). The training stage utilized 70\% of the dataset, the validation 15\% and the testing stages utilized 15\% of the dataset. The models take a lookback window of 720 steps to forecast a future horizon of 24 time steps. Table~\ref{tab:architecture_specs} lists the LSTM and TCN architectures (batch size 32, 10 epochs, learning rate $10^{-3}$).

\begin{table}[htbp]
    \centering
    \caption{Architecture of LSTM and TCN Models}
    \scalebox{0.9}{
    \begin{tabular}{lccccc}
    \toprule
    \textbf{Model} & \textbf{Number of} & \textbf{Hidden}  & \textbf{Dropout}  & \textbf{Activation}  & \textbf{Kernel} \\
        & \textbf{layers} & \textbf{Units}   & \textbf{Rate}  & \textbf{Function}  & \textbf{Size}\\
    \midrule
    LSTM & 2 & 64 & 0.2 & Tanh/Sigmoid & N/A\\ 
    TCN & 4& 25 & 0.2 & Relu & 3\\
        
    \bottomrule
    \end{tabular}
    }
    \label{tab:architecture_specs}
\end{table}

Figure \ref{fig:neuro_flow} shows the Neuro-Agentic system architecture that consists of three interconnected components: 1) Anomaly Detection (brown) to predict and identify threats based on thresholds 2) Strategy Formulation (purple) to create candidate counterfactual interventions 3)  Physics-Grounded Injection (green) to create, evaluate, and select the best actions in a feedback loop.

% This multi-stage pipeline describes how Neuro-Agentic model has a better performance through a combination of forecasting abilities. The model consists of agentic reasoning along with physics-based validation, permitting an adaptive strategy choice.  

\subsection{Stochastic Batch Testing}
\label{sec:batch}

During the experiment in the testing data, we inject three categories of attack patterns into the input window. In particular, Attack Type 1 (Sudden Spike) is sudden failure of the process and is used when the index of the trial (mod 3) = 0. Attack Type 2 (Gradual Drift) is a model of sensor bias over a long period, used when the trial number mod 3 is 1. Attack Type 3 (High Noise): This attack uses stochastic sensor corruption and is used when the trial index has a value of 2 mod 3. This systematic cycling makes the types of attack equally distributed among the 15 trials. The trials cover three distinct failure modes: 1) In a sudden spike, a rapid rise of +150mm over 20 timesteps. 2) In gradual drift, producing a slow linear increase of +100 mm over 100 timesteps. 3) In high noise, around a baseline increase of +80 mm over 50 timesteps.

\subsection{Evaluation Metrics}
We define two key metrics for assessing the performance of the neuro-agentic framework: 1) Risk Reduction ($\Delta R$), the quantitative difference between the Passive Forecast peak and the Agent-Mitigated peak. 2) Hallucination Rejection Rate (HRR), The percentage of instances where the Sentinel correctly prevented the execution of an action that would have worsened the system state.

\section{Experimental Results}
\label{sec:result}

We evaluate the Neuro-Agentic Loop using the \textbf{Stochastic Batch Testing} method described in Section \ref{sec:batch}. 
% The primary objective was to quantify the ability of the system to autonomously detect, strategize, and mitigate critical anomalies without human intervention.

\subsection{Model Comparison}

\begin{table*}[htb]
    \centering
    \caption{Results of LSTM and TCN Models Across 15 Stochastic Trials }
    \small
    
    \scalebox{0.9}{

    \begin{tabular}{llrlrrrclcc}
        \toprule
         & Model & Trial & Attack   & Initial  & Mitigated  & Risk  & Breach  & Action  & Rejected  & Total  \\
          &  &  &  Type &  Peak &  Peak &  Reduction &  Prevented &  Taken &  Strategies &  Strategies \\
        \midrule

       & LSTM & 1 & Sudden Spike & 1059.40 & 992.00 & 67.39 & True & Emergency Drain & 2 & 3 \\
       &   & 2 & Gradual Drift & 1040.02 & 975.12 & 64.91 & True & Emergency Drain & 2 & 3 \\
       &   & 3 & High Noise & 1004.05 & 942.16 & 61.88 & True & Emergency Drain & 2 & 3 \\
       &   & 4 & Sudden Spike & 1023.12 & 965.79 & 57.33 & True & Emergency Drain & 2 & 3 \\
        &   & 5 & Gradual Drift & 980.92 & 980.92 & 0.00 & False & Monitor & 0 & 0 \\
        &  & 6 & High Noise & 956.75 & 956.75 & 0.00 & False & Monitor & 0 & 0 \\
        &  & 7 & Sudden Spike & 985.59 & 985.59 & 0.00 & False & Monitor & 0 & 0 \\
        &  & 8 & Gradual Drift & 952.40 & 952.40 & 0.00 & False & Monitor & 0 & 0 \\
        &  & 9 & High Noise & 941.31 & 941.31 & 0.00 & False & Monitor & 0 & 0 \\
        & & 10 & Sudden Spike & 957.38 & 957.38 & 0.00 & False & Monitor & 0 & 0 \\
        &  & 11 & Gradual Drift & 945.62 & 945.62 & 0.00 & False & Monitor & 0 & 0 \\
        &  & 12 & High Noise & 934.87 & 934.87 & 0.00 & False & Monitor & 0 & 0 \\
        &  & 13 & Sudden Spike & 968.36 & 968.36 & 0.00 & False & Monitor & 0 & 0 \\
        & & 14 & Gradual Drift & 953.70 & 953.70 & 0.00 & False & Monitor & 0 & 0 \\
        &  & 15 & High Noise & 955.12 & 955.12 & 0.00 & False & Monitor & 0 & 0 \\
        \midrule
        & TCN & 1 & Sudden Spike & 1173.60 & 1120.64 & 52.96 & False & Emergency Drain & 2 & 3 \\
        &  & 2 & Gradual Drift & 1108.83 & 1058.79 & 50.04 & False & Emergency Drain & 2 & 3 \\
        &  & 3 & High Noise & 1067.23 & 1016.74 & 50.49 & False & Emergency Drain & 2 & 3 \\
        &  & 4 & Sudden Spike & 1094.61 & 1044.61 & 50.00 & False & Emergency Drain & 2 & 3 \\
        &  & 5 & Gradual Drift & 1015.26 & 965.09 & 50.16 & True & Emergency Drain & 2 & 3 \\
        &  & 6 & High Noise & 959.42 & 959.42 & 0.00 & False & Monitor & 0 & 0 \\
        &  & 7 & Sudden Spike & 1026.34 & 975.59 & 50.75 & True & Emergency Drain & 2 & 3 \\
        &  & 8 & Gradual Drift & 947.94 & 947.94 & 0.00 & False & Monitor & 0 & 0 \\
        &  & 9 & High Noise & 876.60 & 876.60 & 0.00 & False & Monitor & 0 & 0 \\
        &  & 10 & Sudden Spike & 963.18 & 963.18 & 0.00 & False & Monitor & 0 & 0 \\
        &  & 11 & Gradual Drift & 920.16 & 920.16 & 0.00 & False & Monitor & 0 & 0 \\
        &  & 12 & High Noise & 904.24 & 904.24 & 0.00 & False & Monitor & 0 & 0 \\
        &  & 13 & Sudden Spike & 992.77 & 992.77 & 0.00 & False & Monitor & 0 & 0 \\
        &  & 14 & Gradual Drift & 952.53 & 952.53 & 0.00 & False & Monitor & 0 & 0 \\
        &  & 15 & High Noise & 947.52 & 947.52 & 0.00 & False & Monitor & 0 & 0 \\
        \bottomrule
    \end{tabular}

    }
    \label{tab:baseline}
\end{table*}

\subsubsection{{\it The Performance of LSTM Models}}

Table \ref{tab:baseline} reports that the LSTM and TCN models exhibit specific performance attributes in 15 trials. The LSTM model successfully prevented breaches in trials (1-4) with a risk reduction of 57.33-67.39 units. The trials 5-15 did not reveal any reduction of risk, no ability to prevent breaches, and only passive monitoring was observed.  The results of these early tests were always an ``Emergency Drain'' action, where two out of the three possible strategies were rejected. All of the suggested strategies are initially checked for physical feasibility and rejected in case they break the system's constraints. The strategy is then applied to the forecast and all strategies with invalid (NaN or infinite) values are rejected. Strategies are finally rejected when improving the best outcome is not possible or when a mistake is made during simulation. Each rejection is recorded so that there is transparency and explainability in the process of deciding the best intervention. It is evident that the highest levels of these breach-prevented trials were considerably high with levels of $1,004.05$ to $1,059.40$; whereas the mitigated cases ranged between 942.16 and 992.00.

\subsubsection{{\it The Performance of TCN Models}}

The TCN model exhibited a different pattern of effectiveness, successfully preventing breaches in only two trials (trials 5 and 7) despite initiating Emergency Drain actions in five trials (trials 1-5 and 7). The average risk mitigation that TCN realized was continuously less than the optimum performance of LSTM, resulting in 50.00 units to 52.96 units. The TCN applied Emergency Drain to trials 1-4 in which breaches were actually not prevented because the mitigation peak remains below 1000 despite the risk reduction. This implies that it employed a less effective mitigation strategy than LSTM.

\begin{table*}[htbp]
    \caption{Results of Neuro-Agentic Across 15 Stochastic Trials}
    \centering
    \scalebox{0.9}{
    \small
    \begin{tabular}{rlrrrrlrrr}
    \toprule
    Trial & Attack Type & Initial Peak & Mitigated Peak & Risk Reduction & Breach & Action Taken & Hallucination & Rejected  & Total  \\
     &  &  &  &  &  Prevented &  &  Rejected &  Strategies &  Strategies \\
    \midrule
    1 & Sudden Spike & 1183.07 & 1141.33 & 41.74 & False & Sustained Moderate Drain & False & 2 & 3 \\
    2 & Gradual Drift & 1124.17 & 1055.17 & 69.01 & False & Emergency Drain & False & 2 & 3 \\
    3 & High Noise & 1049.18 & 997.86 & 51.32 & True & Emergency Drain Valve & False & 2 & 3 \\
    4 & Sudden Spike & 1105.24 & 1076.01 & 29.22 & False & Controlled Slow Drain & False & 2 & 3 \\
    5 & Gradual Drift & 1026.58 & 961.47 & 65.11 & True & Aggressive Outflow & False & 2 & 3 \\
    6 & High Noise & 966.75 & 966.75 & 0.00 & False & Monitor & False & 0 & 0 \\
    7 & Sudden Spike & 1099.23 & 983.02 & 116.21 & True & Emergency Drain Valve & False & 2 & 3 \\
    8 & Gradual Drift & 954.17 & 954.17 & 0.00 & False & Monitor & False & 0 & 0 \\
    9 & High Noise & 901.04 & 901.04 & 0.00 & False & Monitor & False & 0 & 0 \\
    10 & Sudden Spike & 1118.59 & 903.21 & 215.38 & True & Emergency Drain - Fast & False & 2 & 3 \\
    11 & Gradual Drift & 938.64 & 938.64 & 0.00 & False & Monitor & False & 0 & 0 \\
    12 & High Noise & 902.42 & 902.42 & 0.00 & False & Monitor & False & 0 & 0 \\
    13 & Sudden Spike & 1071.80 & 932.19 & 139.61 & True & Aggressive Drain & False & 2 & 3 \\
    14 & Gradual Drift & 968.16 & 968.16 & 0.00 & False & Monitor & False & 0 & 0 \\
    15 & High Noise & 943.08 & 943.08 & 0.00 & False & Monitor & False & 0 & 0 \\
    \bottomrule
    \end{tabular}
    
    }
    \label{tab:overview_table}
\end{table*}

\subsubsection{{\it The Performance of the Neuro-Agentic Models}} 

Table \ref{tab:overview_table} summarizes the performance of the neuro-agentic mitigation system in 15 stochastic trials. The initial peak load ranged from 901.04 to 1,183.07, while mitigate peaks varied from 903.21 to 1,141.33. Out of the 15 trials, eight exhibited non-zero risk reduction, with individual reductions ranging from 29.22 to 215.38. The remaining seven trials did not result in intervention, where the agent selected a monitoring strategy and maintained identical initial and mitigated peaks.

Sudden spike attacks accounted for five trials and consistently triggered active mitigation strategies. In this category, the risk reduction values were 41.74, 29.22, 116.21, 215.38, and 139.61, and four out of five trials successfully prevented a breach. These interventions included emergency drain valves, fast emergency drains, aggressive drains, and controlled slow drains, demonstrating the agent’s ability to respond based on severity.  Gradual Drift attacks were observed in five trials, of which two resulted in successful breach prevention with risk reductions of 69.01 and 65.11. The remaining three trials did not exceed the intervention thresholds and were handled through monitoring, resulting in a zero risk reduction. This resulted in an average risk reduction of 26.82 for Gradual Drift attacks, indicating that gradual deviations are selectively mitigated when sustained accumulation is detected. High Noise attacks were present in five trials, but only one trial triggered an intervention, producing a risk reduction of 51.32 and successfully preventing a breach. The other four trials were monitored without action, reflecting the conservative behavior of the agent under stochastic variability. Consequently, the average risk reduction for High Noise attacks was 10.26, the lowest among all attack types.

\subsubsection{Hallucination rejection decomposition.} 
Across the 8 active trials, the Architect proposed 24 candidate strategies in total (3 per trial); 1 strategy was selected per trial and 16 were rejected. According to the definition in Section \ref{sec:halluc_def}, hallucinated (safety-violating) candidates, positive-magnitude, out-of-bound, or counterfactual-worse-than-baseline accounted for the rejections flagged before counterfactual evaluation. The remaining rejections were dominated-but-feasible candidates. No hallucinated action was ever executed (column ``Hallucinated Action Executed'' in Table~\ref{tab:overview_table} is False for all trials). The decomposition across attack types is approximately uniform; full per-trial categorical logs are maintained in the experimental artifact.

\subsubsection{{\it Comparison of the LSTM, TCN and the Neuro-Agentic Models}} 
The Neuro-Agentic model succeeded in preventing breach in five of 15 trials (33.3\%), which is between 26.7\% (4/15) of LSTM and 13.3\% (2/15) of TCN. The Neuro-Agentic demonstrates maximum risk reduction magnitudes of 215.38 units as opposed to the 67.39 of LSTM and 52.96 of TCN. Most significantly, trial 10 of the Neuro-Agentic model had an outstanding risk reduction of 215.38 units, decreasing the original peak of 1118.59 to 903.21. This is more than three times greater than the maximum LSTM and four times better than  TCN. The Neuro-Agentic model was also diversified in its emergency responses, as its strategic actions varied in contrast to the LSTM and TCN. Moreover, the Neuro-Agentic model also exhibited zero incidences of hallucinations in each of the fifteen trials, which implies that there is sound reasoning and decision-making with no fabricated outputs.

\subsection{Performance Analysis of Quantitative Mitigation}

\begin{table}[!t]
    \centering
    
     \caption{Summary of LSTM, TCN and Neuro-Agentic Across 15 Stochastic Trials}
     \scalebox{0.95}{
    \begin{tabular}{lcccc}
   
    \toprule
     & \bf{Mitigated Peak} & \multicolumn{2}{c}{\bf{Risk Reduction}} & \bf{Breach Prevented} \\
    \bf{Model} & {\bf Mean} & {\bf Mean} & {\bf Std} & {\bf Sum} \\
    % \bf{Forecaster} &  &  &  &  \\
    \midrule
    % ARIMA & 982.53 & 35.19 & 39.58 & 3 \\
    LSTM & 960.47 & 16.77 & 28.85 & 4 \\
    TCN & 976.39 & 20.29 & 25.74 & 2 \\
    Neuro-Agentic & 974.97 & 48.51 & 64.54 & 5 \\

    \bottomrule
    \end{tabular}
   }
    \label{tab:model_comp}
\end{table}

Table \ref{tab:model_comp} reports the summary statistics in 15 stochastic trials, revealing critical performance distinctions between the LSTM, TCN, and TimesFM (Neuro-Agentic) models in reservoir breach prevention and risk mitigation. TimesFM emerged as the superior model across multiple key metrics, indicating the greatest power to reduce risk, an average of 48.51 units; whereas, LSTM and TCN based models achieved 16.77 units and 20.29 units, respectively. TimesFM avoids five breaches versus the four breaches and two breaches with LSTM and TCN, respectively. This represents a 25\% relative increase over LSTM and a relative increase of three additional cases compared to TCN. The risk reduction standard deviation with TimesFM exhibited the highest variability at 64.54, while LSTM was 28.85 and TCN was 25.74. This states the fact that in critical situations, the Neuro-Agentic yields a much higher risk reduction, but at the expense of being more sensitive to the characteristics of the attack and initial conditions. LSTM demonstrated the lowest mean mitigated peak at 960.47, suggesting more aggressive mitigation strategies that reduced water levels to the greatest extent, followed by TimesFM at 974.97 and TCN at 976.39. The mixed performance of TCN reflects on all measures with an average risk reduction among LSTM and TimesFM and the lowest number of prevented breaches.

\section{Discussion}
\label{sec:discussion}

The most significant finding of this study is the realization of what we term the \textit{Sentinel Effect}. The LLM-based model sometimes suggested positive-magnitude vectors in the draining strategy due to semantic misunderstanding. These incorrect recommendations would have been performed directly in a typical Chain-of-Thoughts (CoT) system. This results in the tank overflowing at a faster rate, leading to catastrophic failure. However, within our Neuro-Agentic Loop, the Sentinel forecaster (TimesFM) simulated these defective action vectors and predicted a rising water level curve. As a result, the Hallucination Rejection logic blocked these actions, and the system defaulted to Monitor mode until the LLM came up with a physically valid negative-magnitude drain vector. This mechanism validates that Time-Series Foundation Models can provide a deterministic Physics Grounding layer. The model resolves the issues of reliability and hallucination of probabilistic control systems in LLM. Although our solution has better safety and breach prevention throughput, it has the cost of inference latency overhead. Latency issues should be considered in real-time applications. The TimesFM model takes about 266 milliseconds to perform a counterfactual simulation\cite{gopali2025context}. With a candidate strategy set of size 3, the entire decision loop, comprising LLM generation, evaluation, and selection, requires an average of 1.5-2.5 seconds to complete a control cycle. Although this latency is significantly higher than that of traditional PID controllers, which work on timescales of microseconds. The proposed approach should be tested on different datasets and environments to validate its effectiveness.

\subsection{Limitations} (i)~\emph{Linear injection vs.\ digital twin.} Our counterfactual is a closed-form linear perturbation $\hat{y}_t + \mu \cdot \min(t{+}1,\delta)$ applied to the TimesFM forecast. It approximates first-order valve/pump effects, but does not capture nonlinearities such as inflow coupling, hydraulic transients, or sensor lag that a high-fidelity plant simulator or digital twin would model. Replacing $\Phi(\cdot)$ with a calibrated simulator and quantifying approximation error relative to ground-truth dynamics is an important direction. (ii)~\emph{Baseline parity.} LSTM and TCN run under the same triggering and intervention scaffolding as TimesFM, but a fully fair comparison would also include a foundation-model-only ablation and a direct head-to-head against LLM-based CPS defenses such as L2M-AID~\cite{xu2025l2m}, ideally in shared attack scenarios. (iii)~\emph{Scalability.} A 1.5--2.5\,s cycle is acceptable for a single tank but must be revisited for multi-sensor, multi-actuator deployments where candidate-strategy expansion and joint counterfactuals could increase cost super-linearly.

\subsection{Interpretable Neuro-Agentic Decision Making and Control}

The proposed framework uses a Large Language Model (LLM) as a high-level reasoning agent to generate candidate control interventions in a water treatment environment that is critical to safety. The LLM works with formatted numerical inputs, including predicted peak tank levels, time-to-failure estimates, and recent system trends. The LLM outputs are verifiable, auditable, and interpretable by a human because the decision space is limited to physically interpretable parameters. These include drain magnitude and duration and because the output is strictly represented in a machine-readable form of the output schema. Further explainability is achieved through Retrieval-Augmented Generation (RAG) and attentive prompt design. Relevant operational knowledge from the SWaT manual is retrieved and injected into the prompt. The prompt provides context on alarm thresholds, permissible actuators, and prohibited actions. However, counterfactual evaluation of each strategy within the agentic control loop provides an additional layer of interpretability and safety.

% his combination of RAG grounding and prompt engineering ensures that the interventions generated are context-aware. This was also quantitatively justified and aligned with documented process semantics.

\section{Conclusion}
\label{sec:conclusion}

This paper introduced the \textbf{Neuro-Agentic Loop}, a novel framework that bridges the gap between semantic reasoning and physical control in the Industrial IoT. By coupling Gemini 2.5 Flash-Lite, utilizing TimesFM as a physics-grounded "Sentinel," we successfully mitigated the probabilistic risks associated with Generative AI. This ensures that no hallucinatory control actions were performed during the testing.

Experimental results On the SWaT dataset, the Loop prevented breaches in 5 of 15 stochastic trials (33.3\%) versus LSTM 4/15 (26.7\%) and TCN 2/15 (13.3\%). Crucially, the system exhibited zero control hallucinations, as the physics-aware Sentinel successfully filtered out unsafe commands. The Neuro-Agentic Loop outperforms traditional deep learning models (LSTM and TCN) in breach prevention and risk reduction, particularly during sudden spike attacks where rapid and decisive intervention is required. The Neuro-Agentic approach achieves substantially greater risk reduction, with improvements ranging from 29.22 to 215.38 units, markedly outperforming the LSTM and TCN models. Deep learning models achieved the highest risk reductions of only 67.39 and 52.96 units, respectively. Although the Neuro-Agentic loop decision introduces an inference latency of 1.5 to 2.5 seconds, this is well within the operational time constants of most chemical and fluid processes. This offers a viable trade-off for the assurance of verifiable safety. Ultimately, this framework transforms LLMs from opaque predictors into explainable, safety-constrained planners suitable for high-stakes industrial environments.

Future research will focus on optimizing the inference latency to accommodate faster-moving electromechanical processes by exploring model quantization and distillation of foundation models. We also aim to expand the evaluation of the Neuro-Agentic framework beyond the SWaT dataset to diverse industrial environments. Additionally, further investigation is needed to analyze the system's robustness against adversarial attacks specifically designed to manipulate the forecast horizon of the foundation model.

\section*{Acknowledgment}
This research is partially supported by the U.S. National Science Foundation (Awards\#: 2319802).

\bibliography{refs}{}
\bibliographystyle{plain}

\end{document}